\documentclass[preprint,12pt]{elsarticle}

\usepackage{amssymb}
\usepackage{amsmath}
\usepackage{booktabs}
\usepackage{url}
\usepackage{graphicx}
\usepackage{subcaption}
\usepackage{multirow}
\usepackage{array}

\journal{Medical Image Analysis}

\begin{document}

\begin{frontmatter}



\title{A fully automated deep-learning pipeline for volumetric endolymphatic hydrops quantification in MRI}


%% =========================
%% Authors
%% =========================
\author[a,b,1]{Caterina Fuster-Barcel\'o}
\ead{cafuster@pa.uc3m.es}
\author[a]{Claudia Castrill\'on}
\ead{100450756@alumnos.uc3m.es}
\author[a]{Laura Rodrigo-Mu\~noz}
\ead{100450920@alumnos.uc3m.es}
\author[c]{Victor Manuel Su\'arez-Vega}
\ead{vvega@unav.es}
\author[d]{Nicol\'as P\'erez-Fern\'andez}
\ead{nperezfer@unav.es}
\author[e]{Gorka Bastarrika}
\ead{bastarrika@unav.es}
\author[a,b,2]{Arrate Mu\~noz-Barrutia}
\ead{mamunozb@ing.uc3m.es}

%% =========================
%% Affiliations
%% =========================
\affiliation[a]{
  organization={Bioengineering Department, Universidad Carlos III de Madrid},
  addressline={Avenida de la Universidad, 30},
  city={Leganes},
  postcode={28911},
  state={Madrid},
  country={Spain}
}

\affiliation[b]{
  organization={Bioengineering Division, Instituto de Investigacion Sanitaria Gregorio Maranon},
  addressline={Calle Ibiza, 43},
  city={Madrid},
  postcode={Es28009},
  state={Madrid},
  country={Spain}
}

%\fntext[label2]{}
\fntext[1]{
  BioVisionCenter, Department of Molecular Life Sciences, University of Zurich,Y55K56, Winterthurerstrasse 190, Zurich, CH-8057, Zürich, Switzerland
}

\affiliation[c]{
  organization={Radiology Department, Clínica Universidad de Navarra},
  addressline={Calle Marquesado de Santa Marta, 1},
  city={Madrid},
  postcode={ES28027},
  state={Madrid},
  country={Spain}
}

\affiliation[d]{
  organization={Otorhinolaryngology Department, Clínica Universidad de Navarra},
  addressline={Calle Marquesado de Santa Marta, 1},
  city={Madrid},
  postcode={ES28027},
  state={Madrid},
  country={Spain}
}

\affiliation[e]{
  organization={Radiology Department, Clínica Universidad de Navarra},
  addressline={Avenida de Pio xII, 36},
  city={Pamplona},
  postcode={ES31008},
  state={Navarra},
  country={Spain}
}

\fntext[2]{
  Neuroscience and Life Sciences Department, Universidad Carlos III de Madrid, Calle de Madrid, 126, Getafe, ES28903, Madrid, Spain
}

\cortext[3]{Corresponding author: Arrate Muñoz-Barrutia}

%% =========================
%% Placeholders for Claudia & Laura (commented so they don't print yet)
%% =========================
% \affiliation[d]{%
%   organization={Affiliation for Claudia Castrillon},
%   addressline={To be completed},
%   city={},
%   postcode={},
%   state={},
%   country={}
% }
%
% \affiliation[e]{%
%   organization={Affiliation for Laura Rodrigo Munoz},
%   addressline={To be completed},
%   city={},
%   postcode={},
%   state={},
%   country={}
% }

\bigskip
%{\small\textsuperscript{*}These authors contributed equally to this work.}

%% Abstract
\begin{abstract}
We present OREHAS (Optimized Recognition \& Evaluation of volumetric Hydrops in the Auditory System), the first fully automatic pipeline for volumetric quantification of endolymphatic hydrops (EH) from routine 3D-SPACE-MRC and 3D-REAL-IR MRI. The system integrates three components—slice classification, inner-ear localization, and sequence-specific segmentation—into a single workflow that computes per-ear endolymphatic-to-vestibular volume ratios (ELR) directly from whole MRI volumes, eliminating the need for manual intervention.

Trained with only 3–6 annotated slices per patient, OREHAS generalized effectively to full 3D volumes, achieving Dice scores of 0.90 for SPACE-MRC and 0.75 for REAL-IR. In an external validation cohort with complete manual annotations, OREHAS closely matched expert ground truth (VSI = 74.3\%) and substantially outperformed the clinical syngo.via software (VSI = 42.5\%), which tended to overestimate endolymphatic volumes. Across 19 test patients, vestibular measurements from OREHAS were consistent with syngo.via, while endolymphatic volumes were systematically smaller and more physiologically realistic.

These results show that reliable and reproducible EH quantification can be achieved from standard MRI using limited supervision. By combining efficient deep-learning–based segmentation with a clinically aligned volumetric workflow, OREHAS reduces operator dependence, ensures methodological consistency. Besides, the results are compatible with established imaging protocols. The approach provides a robust foundation for large-scale studies and for recalibrating clinical diagnostic thresholds based on accurate volumetric measurements of the inner ear.
\end{abstract}


%\item Fully automated pipeline for volumetric hydrops quantification from MRI
%\item Integrates CNN, YOLOv5, and U-Net for inner-ear detection and segmentation
%\item Produces accurate and reproducible endolymph-to-vestibule volume ratios
%\item Outperforms commercial software, achieving  86.1\% and 74.3\% similarity to the manual ground truth
%\item Open-source tool supporting faster and reproducible clinical workflow

%% Keywords

\end{frontmatter}



\section{Introduction}
\label{sec:introduction}
Meniere’s Disease (MD) is a chronic disorder of the inner ear that significantly affects patients’ quality of life through recurrent episodes of vertigo, tinnitus, fluctuating hearing loss, and a sensation of fullness in the affected ear~\cite{bronstein2000}. Despite its clinical relevance, the cause of MD remains unknown~\cite{masoud}. The structural manifestation associated with this disease is characterized by an increase in the volume of endolymph, resulting in distention of the endolymphatic spaces~\cite{buki}. Endolymph is the fluid that fills the cochlea, vestibule, and semicircular canals; its abnormal buildup is known as Endolymphatic Hydrops (EH) and is recognized as a pathological hallmark of MD. In particular, vestibular EH is a specific predictor of definite MD~\cite{Yoshida}. However, evidence of EH has traditionally been verified only through post-mortem histological examinations~\cite{Grkov}, complicating diagnosis during a patient’s lifetime. The underlying cause of EH remains conjectural, with proposed mechanisms including obstruction of endolymphatic flow, osmotic dysregulation, or increased active ions within the endolymph~\cite{Takeda}.

Unilateral manifestation of EH is the most common initial presentation of MD; nevertheless, even in unilateral MD the contralateral ear can show EH within pathological ranges, and the disease may evolve to bilateral involvement over time~\cite{paper_medicos}. Accordingly, imaging and assessing both ears provides insight into the affected side and enables early identification of potential bilateral progression~\cite{evolution}.

For clinical imaging, two MRI sequences are commonly employed in patients with suspected MD. The first is the 3D Relaxation-Enhanced Acquisition Loop with Inversion Recovery (3D-REAL-IR), which provides high-resolution visualization of endolymphatic and perilymphatic spaces after contrast administration and permits accurate volumetric assessment. The second is a native high-resolution cisternographic T2-weighted sequence—referred to here as 3D Sampling Perfection with Application-optimized Contrasts using different flip angle Evolutions Magnetic Resonance Cisternography (3D-SPACE-MRC)—that depicts the fluid-filled inner-ear anatomy without contrast. Used together, these modalities enable complementary structural and functional assessment of the vestibular system~\cite{deng2022comparison}.

At present, the Endolymphatic Ratio (ELR) can be quantified and graded directly from MRI in clinical practice~\cite{paper_medicos}. Despite this advance, EH quantification remains time-consuming because segmentation is performed manually~\cite{Naganawa}. Clinicians must delineate the vestibule boundaries slice-by-slice, a process that is inefficient and susceptible to variability. This motivates the development of automated tools that simplify and standardize the workflow across patients, enabling accurate, real-time ELR calculation while reducing subjectivity.

Recent progress in computer vision and artificial intelligence offers a path toward such automation, particularly via segmentation algorithms. Anatomical features and Regions of Interest (ROI) can be delineated by precise outlining~\cite{RAYED}. Unlike coventional approaches based on thresholding or region growing, deep learning enables semantic segmentation by learning spatial and contextual relationships to assign anatomically meaningful labels to each pixel~\cite{Aakanksha}. This data driven formulation is particularly advantageous in medical imaging, where segmentation accuracy directly influences downstream quantitative analysis and clinical interpretation.

In this paper, we address the limitations of manual EH quantification by introducing a fully automatic, open-source pipeline designed for routine 3D-SPACE-MRC and 3D-REAL-IR MRI stacks, OREHAS (Optimized Recognition \& Evaluation of volumetric Hydrops in the Auditory System). Our approach makes each stage explicit: the pipeline first filters slices to retain only those containing inner-ear anatomy, then localizes the vestibular region, performs sequence-specific segmentation (vestibular space in 3D-SPACE-MRC and vestibular endolymph in 3D-REAL-IR), and finally computes the volumetric ELR for each ear. By releasing open-source code and explicit processing logic, we establish a scalable and computationally efficient framework in which all processing steps are fully traceable. This transparency enables objective and reproducible assessment of endolymphatic hydrops, providing a robust alternative to proprietary clinical workflows. 

\section{Related Work}

Semi-automatic strategies for quantifying vestibular EH have been explored using 3D-SPACE-MRC and 3D-REAL-IR MRI sequences. Suárez-Vega et al. developed a workflow that required manual segmentation of regions of interest in both sequences (approximately eight slices per ear) and employed the Siemens Syngo.via VB50B platform to estimate total volume from these delineations \cite{paper_medicos}. Although this approach represented an important step toward quantification, it still relied heavily on expert input. Other semi-automatic methods based on thresholding, region growing, or edge detection were proposed, but all required user supervision until the emergence of machine-learning-based segmentation \cite{Vaidyanathan}.

To achieve full automation, deep learning (DL) methods were introduced. Cho et al.~\cite{Cho2020} proposed INHEARIT (INner-ear Hydrops Estimation via ARtificial InTelligence), a VGG-19-based network for segmenting cochlear and vestibular structures from 3D-FLAIR and inversion-recovery MRI, estimating the ELR rom surface measurements with strong agreement to expert annotations. Convolutional architectures such as U-Net and its variants have since become standard for inner-ear segmentation \cite{Vaidyanathan, Park2021, Yoo}. Park et al.~\cite{Park2021} later introduced INHEARIT-v2, which combined Inception-v3 and U-Net to compute ELR from multiple MRI contrasts, again relying on surface-based measurements. More recently, we proposed a U-Net with a ResNet34 encoder for estimating area-based ELR from 3D-SPACE-MRC and 3D-REAL-IR~\cite{caseib}. However, these methods operate on selected slices rather than full volumes, limiting their ability to capture the three-dimensional extent of vestibular endolymph~\cite{Yoo}.

To overcome these limitations, several studies have explored 3D convolutional architectures. Vaidyanathan et al.~\cite{Vaidyanathan} developed a 3D U-Net with attention gates for multi-structure segmentation on T2-weighted MRI, achieving high accuracy and generalizability. Similarly, Yoo et al.~\cite{Yoo} applied 3D U-Net to MR cisternography and HYDROPS-Mi2 data to compute volumetric ELR. Although 3D models provide superior spatial context, they require large annotated datasets, extensive memory, and long training times~\cite{2dv3d}. In contrast, 2D U-Net architectures offer a more efficient and scalable alternative by processing individual slices, maintaining competitive accuracy with reduced computational cost~\cite{2dv3d}.

Alternative non-DL methods have also been proposed. Gerb et al. introduced VOLT, a pipeline combining local thresholding and CNN-based region registration to segment the endolymphatic space~\cite{VOLT}. While it avoids large datasets and training demands, VOLT does not compute the ELR and focuses solely on volumetric segmentation.

In summary, recent advances—particularly 3D DL approaches—have improved volumetric ELR estimation but remain computationally intensive and often require extensive supervision~\cite{Yoo}. For clinical deployment, tools must be modular, efficient, and reproducible. Many existing methods compute ELR across multiple inner-ear structures, whereas our focus is on the vestibule, the most clinically relevant region. To address these gaps, we propose an end-to-end pipeline that automates slice selection, localization, and segmentation using a 2D U-Net backbone~\cite{Park2021, Azad}. By aggregating slice-wise predictions into a full 3D reconstruction, our approach yields volumetric ELR estimates consistent with clinical workflows that annotate slice-by-slice~\cite{Naganawa}.

To our knowledge, no prior work has reported a fully automated inner-ear analysis pipeline encompassing both detection and segmentation. Previous studies have focused on isolated segmentation tasks without automated preprocessing. Our framework introduces a CNN-based slice classifier, similar to those used for brain MRI classification–\cite{liang2021alzheimer, sahoo2024alzheimer}, followed by a YOLO-based detector for anatomical localization–\cite{iriawan2024yolo, chen2024enhancing}. This end-to-end design, applied to inner-ear MRI, establishes a foundation for automated, reproducible, and scalable EH quantification.

\section{Materials and Methods}
\label{sec:matandmeth}
\begin{figure}[ht!]
  \centering
  \includegraphics[width=1.1\textwidth]{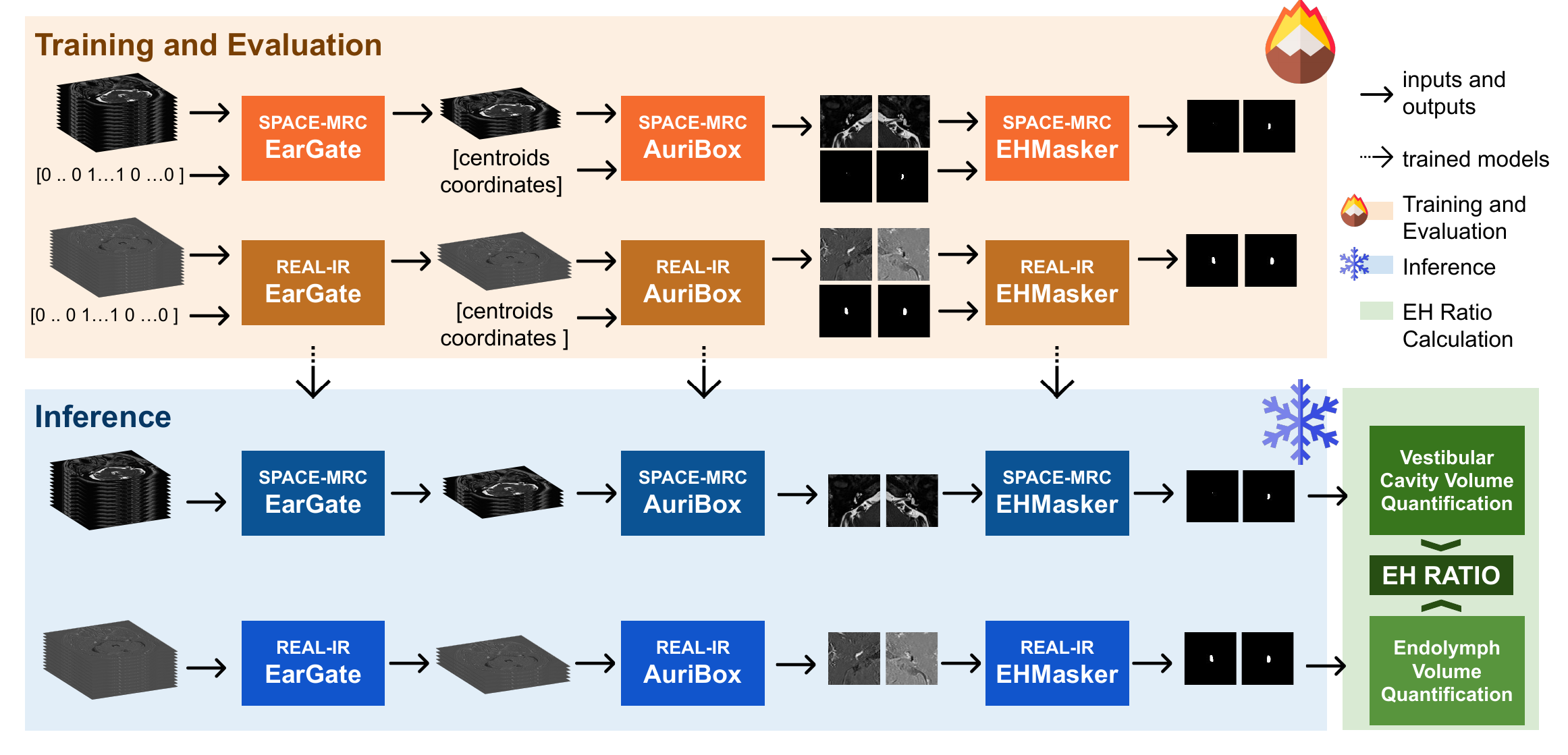}
  \caption{Overview of the EH‐Ratio analysis pipeline. In both training/evaluation (top, peach) and inference (bottom, blue) phases, SPACE‐MRC and REAL‐IR volumes are  sequentially processed by three deep learning modules: (1) \textit{EarGate} performs slice selection to retain inner ear anatomy: (2) \textit{AuriBox} localizes the region of interest through bounding box detection, and (3) \textit{EHMasker} generates segmentation masks for the vestibular and endolymphatic spaces. The reconstructed 3D masks are subsequently used for volumetric quantification of the EH ratio.}
  \label{fig:pipeline}
\end{figure}

This section describes the dataset, processing workflow, and network architectures underlying the fully automated pipeline for EH ratio analysis. The framework processes two volumetric MRI sequences per patient (3D-SPACE-MRC and 3D-REAL-IR) and employs a sequence of deep networks to perform end-to-end EH quantification. During training, annotated data guide model optimizations; in inference, the integrated pipeline operates automatically without manual intervention or ground-truth input, replicating a realistic clinical workflow. The overall architecture, illustrated in Figure \ref{fig:pipeline}, consists of distinct training and evaluation stages followed by an autonomous inference phase that produces the final volumetric EH ratio for each ear. 

\subsection{Data Properties and Acquisition}

\subsubsection{Patient Cohort}
This study includes MRI data from 90 anonymized individuals drawn from the clinical archive of Clínica Universidad de Navarra (CUN). Of these, 83 patients had a definitive diagnosis of unilateral Ménière’s disease and 7 served as control subjects. Ages ranged from 28 to 75 years (mean ± SD: 55 ± 11 years), with a balanced sex distribution (51.11\% male, 48.89\% female). All procedures received local ethics committee approval, and informed consent was obtained for contrast-enhanced MRI acquisition and research use of the data.

\subsubsection{MRI Acquisition Protocol}
All examinations were performed on 3 T Siemens Magnetom scanners (Healthineers, Erlangen, Germany)—either a Magnetom Vida with a 20‐channel coil or a Magnetom Skyra with a 32‐channel coil—using dedicated head/neck arrays. Four hours after intravenous administration of gadolinium‐based contrast agent (Gadovist, 1.0 mmol/mL; dose 0.1 mmol/kg), two volumetric 3D sequences were acquired in succession, for a combined scan time of approximately 16 minutes (Table \ref{tab:acq_params}).

The heavily T2‐weighted 3D‐SPACE‐MRC sequence (see Table \ref{tab:acq_params} and shown in \ref{fig:modalities}) shows contrast between cerebrospinal fluid, perilymphatic fluid, and surrounding bone. In our pipeline, SPACE‐MRC volumes are used to delineate the full vestibular cavity—including the bony labyrinth—and provide the reference mask for total vestibular volume, a key denominator in the EH‐ratio computation.

By contrast, the 3D‐REAL‐IR sequence (also detailed in Table \ref{tab:acq_params} and Figure \ref{fig:modalities}) employs an inversion recovery tuned to suppress perilymphatic (contrast‐enhanced) signal and highlight non‐enhanced endolymph. This configuration produces high‐intensity endolymphatic regions against a dark perilymphatic background, enabling direct volumetric measurement of endolymphatic space. It is noteworthy that the quality of the REAL-IR images is notably inferior than the SPACE-MRC ones resulting in a much more complex analysis.

\begin{table}[ht!]
  \centering
  \caption{Sequence parameters for 3D‐SPACE‐MRC and 3D‐REAL‐IR MRI acquisitions, including voxel size, timing, and acquisition settings.}
  \label{tab:acq_params}
  \small
  \renewcommand{\arraystretch}{1.1}
  \begin{tabular}{@{}lcc@{}}
    \toprule
    \textbf{Parameter}              & \textbf{3D‐SPACE‐MRC}      & \textbf{3D‐REAL‐IR} \\
    \midrule
    Voxel size (mm)                 & 0.5×0.5×0.5             & 0.5×0.5×0.8     \\
    Matrix size (px)                & 320×320                   & 324×384           \\
    Slice thickness (mm)            & 0.5                         & 0.8                 \\
    Repetition time TR (ms)         & 1400                        & 16000               \\
    Echo time TE (ms)               & 155                         & 551                 \\
    Inversion time TI (ms)          & –                           & 2700                \\
    Flip angle (°)                  & 120                         & 125                 \\
    Bandwidth (Hz/px)               & 289                         & 434                 \\
    Number of excitations (NEX)     & 2                           & 2                   \\
    Number of slices                & 56                          & 112                 \\
    Acquisition time (min)          & 5                           & 11                  \\
    \bottomrule
  \end{tabular}
\end{table}

\subsubsection{Annotation and Ground Truth}
\begin{figure}[ht!]
    \centering
    \includegraphics[width=0.8\linewidth]{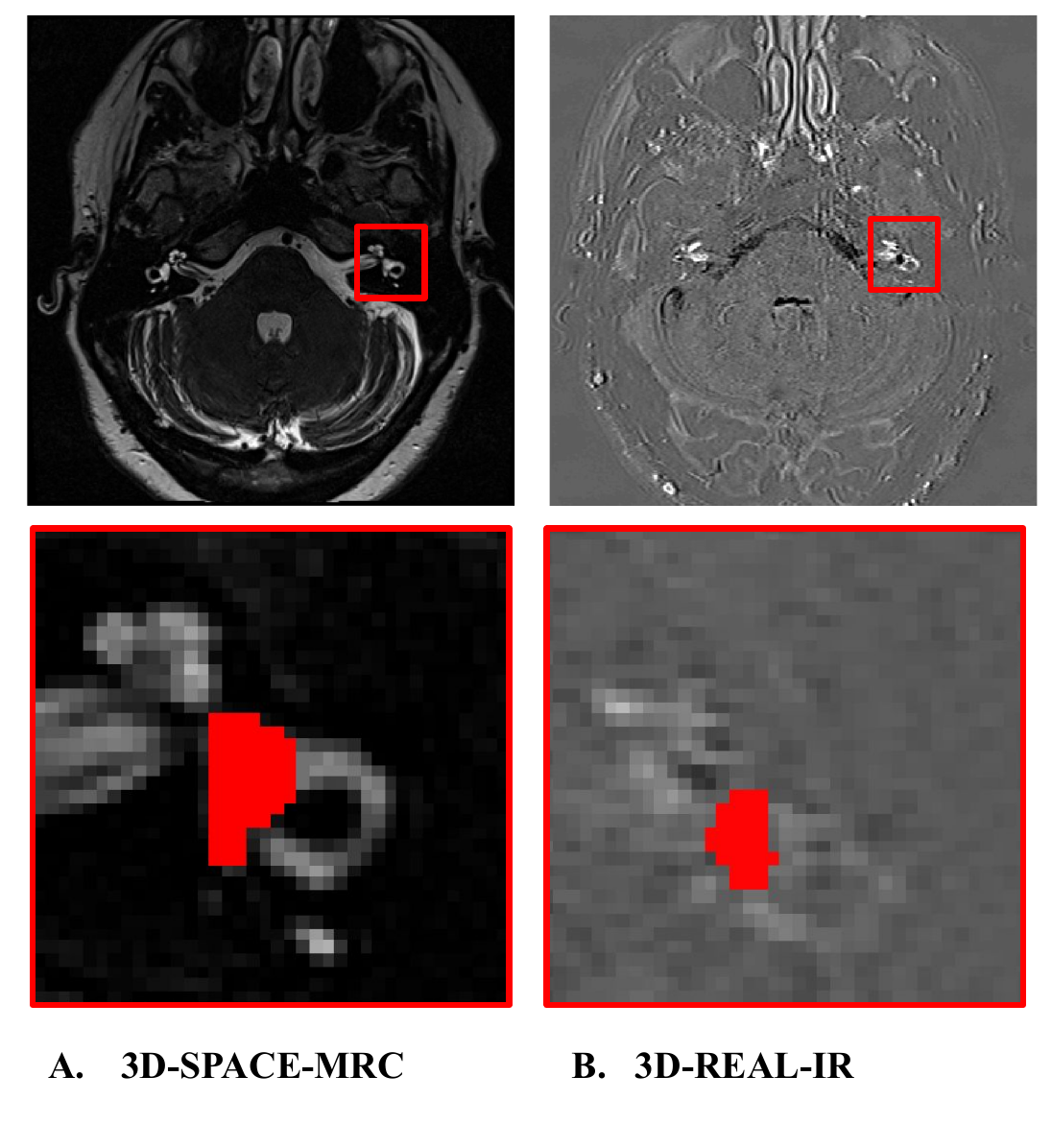}
  \caption{Representative axial slices from a single subject: (a) 3D-SPACE-MRC image showing the delineated vestibular cavity (hyperintense, black contour); (b) 3D-REAL-IR image depicting the endolymphatic compartment (hypointense, white contour).}
  \label{fig:modalities}
\end{figure}

For each patient and imaging sequence, the clinician-selected axial slice (one per volume, see Figure \ref{fig:modalities}) was displayed on the Siemens console where expert neuroradiologists draw rough contours for both the vestibular and endolymphatic spaces. The OpenApps Syngo.Via platform then computed total vestibular and endolymphatic volumes by interpolating across non-annotated slices using a proprietary algorithm and automatically reported the resulting EH ratio. Only the annotated slice, its corresponding mask, and the derived volumetric and ratio values were available for this study, as the interpolation and integration procedures implemented by the software are not publicly disclosed. An example of a clinician-provided contour and its mask overlay is shown in Figure \ref{fig:modalities}.

\subsection{Data Preparation}
All MRI volumes were provided in DICOM format. A custom Python script was used to convert each 3D volume into individual TIFF (\texttt{.tif}) slices, automatically removing corrupted or non-informative frames. Filenames were standardized to include patient identifier, sequence type (SPACE-MRC or REAL-IR), and slice index. The resulting TIFF images were organized into sequence-specific folders with patient-level subdirectories to enable efficient data access. 

Because the pipeline include three stages (classification, detection, and segmentation) each network was supplied with a modality-specific and preprocessed version of the TIFF data optimized for its respective task. 

\subsection{EarGate: slice-level classification}
\label{subsec:eargate}

EarGate is the first stage of the pipeline, responsible for identifying axial slices that contain inner-ear anatomy. Each slice is classified as "ear" or "non-ear" to remove irrelevant regions such as brain, bone, or soft tissue. This preselection step reduces computational load and ensures that subsequent detection and segmentation modules operate only on anatomically relevant data, improving overall accuracy and efficiency. 

\subsubsection{Annotation and Pre-Processing}
Two trained annotators independently reviewed all TIFF slices from the SPACE-MRC and REAL-IR volumes, assigning binary labels ("1" for slices containing inner-ear structures, "0" otherwise). Label consistency was verified by two board-certified neuroradiologists, ensuring accurate identification of ear-containig slices.

To preserve anatomical realism, standard augmentations such as flipping, color jittering, and large affine transformations were omitted. Only minor in-plane rotations $(\pm {10}^{o})$ and slight brightness adjustments, reflecting natural variability in patient position and scanner conditions, were applied during training. 

\subsubsection{Custom FIve-Layer CNN}
The initial EarGate model was a custom five-layer convolutional neural network (CNN) trained from scratch without pretrained weights. Each input slice (3×224×224
pixels) was processed through five convolutional blocks. The first three blocks increased the number of channels from 3→32→64→128, each followed by batch normalization, ReLU activation, 2×2 max pooling, and dropout (p=0.5). The final two blocks expanded channels to 256 and 512, using convolution, batch normalization, and ReLU, but omitted dropout to preserve deeper feature representations. The resulting feature maps were flattened and passed to a fully connected head consisting of a 1,024-unit layer with ReLU and dropout (p=0.5), followed by a two-class output layer for ear versus non-ear classification.

The network was trained separately on SPACE-MRC and REAL-IR data to accommodate modality-specific contrasts: vestibular and perilymphatic structures appear hyperintense in SPACE-MRC, whereas endolymphatic regions are hypointense in REAL-IR. Separate training ensured robust detection across both image types.

\subsubsection{ResNet 50 Fine-Tuning}
To benchmark a state-of-the-art model for slice classification, we fine-tuned ResNet50~\cite{he2016deep}, a 50-layer convolutional network with residual bottlenecks and identity skip connections that support stable training in deep architectures. The model was initialized with ImageNet-pretrained weights, and its 1,000-class fully connected layer was replaced with a compact head comprising dropout (p=0.5) and a linear layer outputting two logits for ear versus non-ear classification.

All network parameters were kept trainable to enable full fine-tuning on MRI data, allowing adaptation of both low- and high-level features to the specific intensity patterns of SPACE-MRC and REAL-IR sequences. The global average pooling layer preceding the classifier produced a 2,048-dimensional embedding capturing modality-specific anatomical representations used for final label prediction.

\subsubsection{Training and Evaluation} 
All classification experiments used patient-level fivefold cross-validation to ensure subject independence between training and validation. Models were trained for up to 50 epochs with early stopping after 10 stagnant epochs. The custom CNN was optimized with Adam, and ResNet50 fine-tuing employed SGD with momentum. Borth used ReduceLROnPlateau scheduler that reduced the learning rate by a factor or 0.1 after five stagnant epochs. 

For SPACE-MRC data, we compared standard cross-entropy, focal loss, and weighted cross-entropy (class weights [1.0, 2.0]); for REAL-IR, we tested weighted cross-entropy, binary cross-entropy with logits, and unweighted cross-entropy. Input slices were resized to 224×224 pixels, intensity-normalized per volume, and augmented with small rotations (±10°) and brightness shifts to reflect scanner variability. Performance was measured on each fold using accuracy, precision, recall, and F1 score, reported as mean values across folds. 

\subsubsection{Classification Post-Processing}
To eliminate anatomically implausible predictions, EarGate outputs were refined using a two-stage smoothing procedure. First, isolated misclassified slices were corrected based on adjacent labels (e.g.\ 0,1,0 → 0,0,0; 1,0,1 → 1,1,1). Second, continuity was enforced by retaining a single continuous "ear" segment per volume, ensuring no internal gaps. This post-processing step reduced spurious detections and stabilized downstream detection and segmentation.  

\subsection{AuriBox: object detection for inner-ear ROIs}
\label{subsec:auribox}

\subsubsection{Annotation procedure: finding centroids}
During training (Figure~\ref{fig:pipeline}), clinician-verified “ear” slices were used instead of EarGate outputs to prevent bias from misclassified inputs. A custom Fiji macro displayed all TIFF slices for each patient, enabling annotators to mark the centroids of the left and right cochlear regions. The recorded (x,y) coordinates were stored in a CSV file with corresponding image identifiers.

A Python script converted these coordinates into axis-aligned bounding boxes centered on each centroid, using fixed dimensions sufficient to encompass the inner-ear anatomy in both SPACE-MRC and REAL-IR modalities. Coordinates and box sizes were normalized by image width and height and saved in YOLO format: \verb|<class> <x_center> <y_center> <width> <height>|

%\begin{quote}
%\texttt{<class>} \texttt{<x\_center>} \texttt{<y\_center>} \texttt{<width>} %\texttt{<height>}
%\end{quote}
Each image was paired with a corresponding “.txt” annotation file, where class identifiers distinguished left from right ears. The full dataset structure—organized by modality and patient ID—was visually verified by board-certified neuroradiologists for anatomical accuracy and labeling consistency.

\subsubsection{YOLOv5 for inner-ear detection}
Inner-ear localization within the preselected “ear” slices was performed using a YOLOv5su detector fine-tuned through the Ultralytics API. YOLOv5su is a lightweight, single-shot model that jointly predicts bounding boxes and class scores in a single forward pass, enabling efficient slice-level inference. The network was initialized with COCO-pretrained weights, modified to include a single “inner-ear” class, and adapted to handle paired SPACE-MRC and REAL-IR inputs.

The architecture combines a CSPDarknet backbone, which enhances gradient flow and feature reuse, with a PANet-style neck that fuses multi-scale representations via top-down and bottom-up pathways. The detection head outputs predictions at three spatial scales, each using three anchor boxes (nine total). During inference, redundant detections were removed using non-maximum suppression with an IoU threshold of 0.45, retaining only the most confident bounding boxes.

\subsubsection{Training Parameters and Evaluation Metrics}
Each YOLOv5su detector was trained and evaluated using patient-level fivefold cross-validation to ensure subject independence between training and validation. An initial three-epoch prototyping phase was used to validate annotations and hyperparameters. Final models were trained for up to 50 epochs using stochastic gradient descent with momentum 0.937, weight decay $5\times10^{-4}$, and an initial learning rate of 0.01, reduced via a \texttt{ReduceLROnPlateau} scheduler when validation loss plateaued. 

Data augmentation was performed on the fly through the Ultralytics API, including Mosaic tiling, random perspective distortion, and random scaling or cropping. During inference, predictions were filtered at a confidence threshold of 0.3 and refined using non-maximum suppression with an IoU of 0.45. Performance was evaluated on held-out folds at an IoU threshold of 0.5, reporting mean average precision (mAP@0.5), precision, and recall, averaged across the five folds.

\subsection{EHMasker: segmentation of the inner-ear structure}
\label{subsec:ehmasker}

\subsubsection{Manual annotation procedure}
Manual ground-truth masks were generated for a limited subset of slices due to the time-intensive nature of contouring. For each patient, three to six axial slices per ear were selected from the DICOM data (converted to 8-bit TIFF): the slice showing the maximal cross-section of the region of interest and one to two adjacent slices anteriorly and posteriorly. As ear symmetry and head positioning varied across subjects, the number of slices per ear differed accordingly.

Expert neuroradiologists at CUN defined the annotation protocol, and contours were drawn using the Labkit plugin in Fiji (ImageJ). Vestibular cavities were delineated on SPACE-MRC images, while endolymphatic compartments were outlined on corresponding REAL-IR slices. The final dataset comprised 437 SPACE-MRC and 409 REAL-IR slices, each paired with a binary mask.

\subsubsection{Image Preprocessing}
Annotated images and corresponding binary masks were augmented by horizontal flipping on training folds, exploiting the bilateral symmetry of the inner ear to increase the sample size without introducing unrealistic anatomy. Each full-slice TIFF was then cropped into two 
patches of size 96×96 pixels centered on the left and right vestibular regions—using manual centroids for training and YOLOv5su detections for validation and testing. This cropping strategy reduced background content, alleviated class imbalance, and standardized input resolution for the U-Net segmenter. Finally, all patches were min–max normalized to the [0,1] range per image, harmonizing intensity distributions across patients and between SPACE-MRC and REAL-IR modalities. These steps yielded a consistent and anatomically realistic dataset for model training and evaluation.

\subsubsection{U-Net Architecture}
The segmentation model follows a standard U-Net encoder-decoder design with symmetric skip connections linking corresponding layers. Each encoder block performs two 3×3 convolutions with padding, followed by normalization, LeakyReLU activation, and 2×2 max pooling, The decoder mirrors this structure, using transposed convolutions for upsampling and skip connections to restore fine spatial detail. A final 1×1 convolution with sigmoid activation outputs pixel-wise probabilities for binary segmentation. 

To enhance stability and generalization on inner-ear MRI data, batch normalization was replaced by Group Normalization (encoder) and instance Normalization (bottleneck) to accommodate small batch sizes. LeakyReLU activations preserved gradient flow, and all layers were initialized using Kaiming initialization. Dropout (p = 0.3) was applied within the bottleneck and selected decoder blocks for regularization. 

\subsubsection{Training Parameters and Evaluation Metrics}
The U-Net was trained using patient-level splits (80\% training, 20\% testing), with 10\% of the training data reserved for validation. Models were trained for up to 100 epochs using Adam (learning rate $1\times10^{-4}$, weight decay $1\times10^{-5}$, batch size 16). A \texttt{ReduceLROnPlateau} scheduler reduced the learning rate by 0.5 after five stagnant epochs, and early stopping (patience 10) prevented overfitting. 

To address class imbalance, five loss formulations were compared: Binary Cross-Entropy (BCE), Focal Loss, Dice Loss, a hybrid BCE+Dice Loss, and Tversky Loss. The hybrid BCE+Dice objective offered the best compromise between stable convergence and overlap accuracy on validation data.

Performance was evaluated on the held-out test set using Dice Similarity Coefficient (DSC), Intersection-over-Union (IoU), and pixel-wise recall: 

\[
\mathrm{DSC} = \frac{2|X \cap Y|}{|X| + |Y|}, \quad
\mathrm{IoU} = \frac{|X \cap Y|}{|X| + |Y| - |X \cap Y|}, \quad
\mathrm{Recall} = \frac{\mathrm{TP}}{\mathrm{TP} + \mathrm{FN}}.
\]

While IoU provides a stricter overlap measure, DSC was prioritized for reporting due to its robustness in small anatomical structures. Recall quantified sensitivity to the full extend of the inner-ear region. All metrics were averaged across test subjects.

\subsubsection{Segmentation Post‐Processing}
During inference, the U-Net produces 2D binary masks for each 96×96-pixel ear patch across all SPACE-MRC and REAL-IR slices. These slice-wise predictions are stacked in axial order to reconstruct a 3D mask for each ear. Volumetric post-processing then removes small isolated components and fills internal gaps, retaining only the largest connected structure to ensure anatomical consistency. The resulting 3D mask is subsequently used to compute the endolymph-to-vestibule ratio (EH-ratio).

During training and validation, where only a few annotated slices are available, segmentation performance is assessed directly on 2D ground-truth masks without volumetric post-processing. 

\subsection{Volumetric ELR Calculation}
Because training and validation include only 3-6 manually annotated slices per ear, hydrops severity is estimated using an area-based endolymph-to-vestibule ratio. For each annotated slice, the endolymphatic area in REAL-IR is divide by the vestibular area in SPACE-MRC to obtain a slice-level ELR. Given that the inner ear spans 10-15 axial slices, this surface-based measure is used solely to monitor segmentation accuracy and not as a volumetric ground truth. 

In the fully automated inference phase, binary segmentation masks are generated for every slice containing ear structures and reassembled in axial order to form 3D volumes of the vestibule and endolymphatic spaces. The total number of foreground voxels in each mask is multiplied by the corresponding voxel size (0.5×0.5×0.5mm³ for SPACE‐MRC; 0.5×0.5×0.8mm³ for REAL‐IR) to compute the true physical volumes. The volumetric Endolymph-to-Vestibule Ratio (ELR) is then defined as
\begin{equation}
\mathrm{ELR}\;[\%] \;=\; \frac{V_{\mathrm{endolymph}}}{V_{\mathrm{vestibule}}}\times 100
\end{equation}
where \(V_{\mathrm{endolymph}}\) and \(V_{\mathrm{vestibule}}\) denote the computed 3D volumes of the endolymphatic space and the total vestibular cavity, respectively.  

In clinical routine, neuroadiologists typically compute the ELR using syngo.via (Siemens Healthineers) platform. After manually outlining the vestibular endolymphatic regions on a small number of representative slices, the software interpolates these contours across unannotated slices using a proprietary algorithm and integrates them to estimate total volumes. Although this approach provides a practical estimate of ELR, it depends on manual input and undisclosed interpolation, which limits reproducibility and may introduce systematic overestimation of endolymphatic volume. In contrast, our pipeline performs this computation directly from the complete 3D segmentation, ensuring full transparency and voxel-level accuracy. 

\subsection{Pipeline Integration \& Runtime Environment} \label{subsec:pipeline-integration}
After model selection through cross-validation, the optimized weights of EarGate, AuriBox, and EHMasker were frozen and integrated into a single end-to-end inference pipeline (Figure~\ref{fig:pipeline}). To prevent data leakage, each module was retrained on the full training cohort (80\%) and evaluated on the remaining 20\%. During inference, the pipeline performs slice classification, ROI detection, segmentation with 3D post-processing, and EH-ratio computation for both SPACE-MRC and REAL-IR volumes, without manual intervention once the input folder is specified.

The workflow is implemented in Python 3.8+, using PyTorch for classification and segmentation, the Ultralytics YOLOv5 API for detection, and standard libraries for preprocessing and post-processing (NumPy, OpenCV, SimpleITK). On a workstation equipped with an NVIDIA RTX 4090 GPU and 128 GB RAM, full processing of 10 patients (both modalities) requires approximately 2 minutes, including model inference and volumetric computations. The source code is available at \url{https://github.com/BSEL-UC3M/EHRatioAnalysis}, ensuring reproducibility and facilitating future deployment as a syngo.via OpenApp, where clinicians can obtain an automated EH-ratio report directly from DICOM input.

For final validation, five additional cases from Clinical Universidad de Navarra, independent from model development, were fully annotated at the slice level to provide voxel-wise ground truth of the entire vestibular volume. These data enabled a real-world evaluation of OREHAS predictions against clinician-verified ELR measurements, addressing inconsistencies observed with the proprietary syngo.via software. 

The modular design of the OREHAS pipeline—comprising EarGate for slice classification, AuriBox for inner-ear localization, and EHMasker for segmentation—was deliberately chosen to maximize interpretability and minimize data leakage. Each module performs a well-defined task on non-overlapping data subsets, ensuring that training, validation, and inference remain fully separated at every stage. This sequential structure allows intermediate outputs to be inspected independently, facilitates module-level debugging and replacement, and provides clear traceability from raw MRI input to final volumetric ELR computation.

\section{Results}
\label{sec:results}

\subsection{EarGate performance}

EarGate classifies MRI slices as ear-containing or non-ear to ensure that only relevant data are processed downstream. Two architectures—a custom 5-layer CNN and a ResNet50—were evaluated using several loss functions (Cross-Entropy, Weighted Cross-Entropy, Multi-Focal, and BCE) on both 3D-SPACE-MRC and 3D-REAL-IR datasets.

Table~\ref{tab:classification} summarizes the results for the REAL-SPACE-MRC dataset. Across all loss functions, the custom CNN consistently outperformed ResNet50.  Weighted Cross-Entropy achieved the best trade-off between precision and recall, yielding an F1 score of $0.85$ and recall of $0.92$.

\begin{table}
\centering
\caption{EarGate slice-classification performance on 3D-SPACE-MRC and 3D-REAL-IR (mean $\pm$ SD over five folds). Bold values indicate the best metric per dataset. }
\label{tab:classification}
\small
\renewcommand{\arraystretch}{1.2}
\setlength{\tabcolsep}{6pt}
\begin{tabular}{l l c c c c}
\toprule
\textbf{Dataset} & \textbf{Model / Loss} & \textbf{Acc. (\%)} & \textbf{F1} & \textbf{Prec.} & \textbf{Rec.} \\
\midrule
\multirow{4}{*}{3D-SPACE-MRC}
 & CNN + CrossEntropy   & \textbf{93.5} $\pm$ 0.8 & 0.84 & 0.87 & 0.81 \\
 & CNN + Multi-Focal    & 93.2 $\pm$ 0.9 & 0.83 & \textbf{0.89} & 0.78 \\
 & CNN + Weighted CE    & 93.3 $\pm$ 2.0 & \textbf{0.85} & 0.80 & \textbf{0.92} \\
 & ResNet50 + CE        & 90.0 $\pm$ 1.1 & 0.75 & 0.80 & 0.70 \\
\midrule
\multirow{4}{*}{3D-REAL-IR}
 & CNN + Weighted CE    & 95.9 $\pm$ 0.4 & 0.57 & 0.67 & 0.49 \\
 & CNN + BCE            & \textbf{96.0} $\pm$ 0.4 & \textbf{0.59} & \textbf{0.71} & 0.51 \\
 & CNN + CE             & 95.8 $\pm$ 0.4 & 0.57 & 0.67 & 0.49 \\
 & ResNet50 + CE        & 95.6 $\pm$ 0.2 & 0.58 & 0.63 & \textbf{0.54} \\
\bottomrule
\end{tabular}
\end{table}

Performance on the 3D-REAL-IR dataset Table~\ref{tab:classification} was more challenging due to strong class imbalance. Accuracy remained high (approximately 96\%), but F1 scores dropped to $0.57-0.59$, reflecting reduced sensitivity to the minority (ear) class. Among the CNN models, BCE and Weighted Cross-Entropy performed best, while ResNet50 offered no clear advantage.  

Overall, the lightweight CNN achieved the best performance on SPACE-MRC and comparable accuracy to ResNet50 on REAL-IR, with substantially higher efficiency. Weighted loss functions improved recall under class imbalance, confirming the suitability of the custom CNN for this stage of the pipeline. 

\subsection{AuriBox performance}

As described in Section \ref{subsec:auribox}, AuriBox uses YOLOv5su detectors trained separately on SPACE-MRC and REAL-IR data to localize the left and right inner ears within MRI slices.

Both models achieved near-perfect detection accuracy (Table~\ref{tab:auribox_results}). For SPACE-MRC, two right-ear instances were missed; for REAL-IR, one. These rare false negatives had no effect on downstream analysis, since subsequent steps rely only on correctly detected bounding boxes. 

\begin{table}
\centering
\caption{AuriBox detector (YOLOv5su) detection results for 3D-SPACE-MRC and 3D-REAL-IR test sets, reporting accuracy and counts of true positives (TP) and false negatives (FN).}
\label{tab:auribox_results}
\small
\renewcommand{\arraystretch}{1.2}
\setlength{\tabcolsep}{6pt}
\begin{tabular}{l l c c c}
\toprule
\textbf{Dataset} & \textbf{Class} & \textbf{Accuracy (\%)} & \textbf{TP} & \textbf{FN} \\
\midrule
\multirow{2}{*}{3D-SPACE-MRC} 
 & Left ear  & 100.0 & 148 & 0 \\
 & Right ear & 98.7  & 148 & 2 \\
\midrule
\multirow{2}{*}{3D-REAL-IR}   
 & Left ear  & 100.0 & 90  & 0 \\
 & Right ear & 98.9  & 90  & 1 \\
\bottomrule
\end{tabular}
\end{table}

\begin{figure}[ht!]
    \centering
    \begin{subfigure}{0.48\linewidth}
        \centering
        \includegraphics[width=\linewidth]{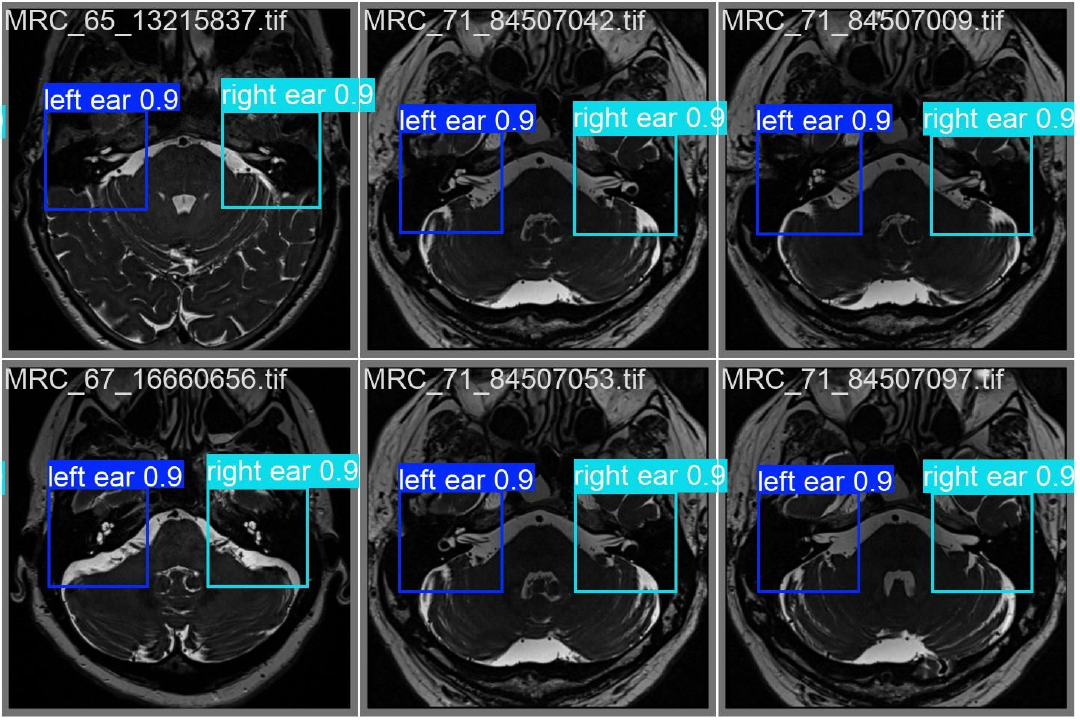}
        %\caption{Representative YOLOv5su detections on the 3D-SPACE-MRC dataset. Predicted bounding boxes for the \textit{left ear} (blue) and \textit{right ear} (cyan) are shown with confidence scores.}
        \label{fig:mrc_test_preds}
    \end{subfigure}
    \hfill
    % Right image: REAL-IR
    \begin{subfigure}{0.48\linewidth}
        \centering
        \includegraphics[width=\linewidth]{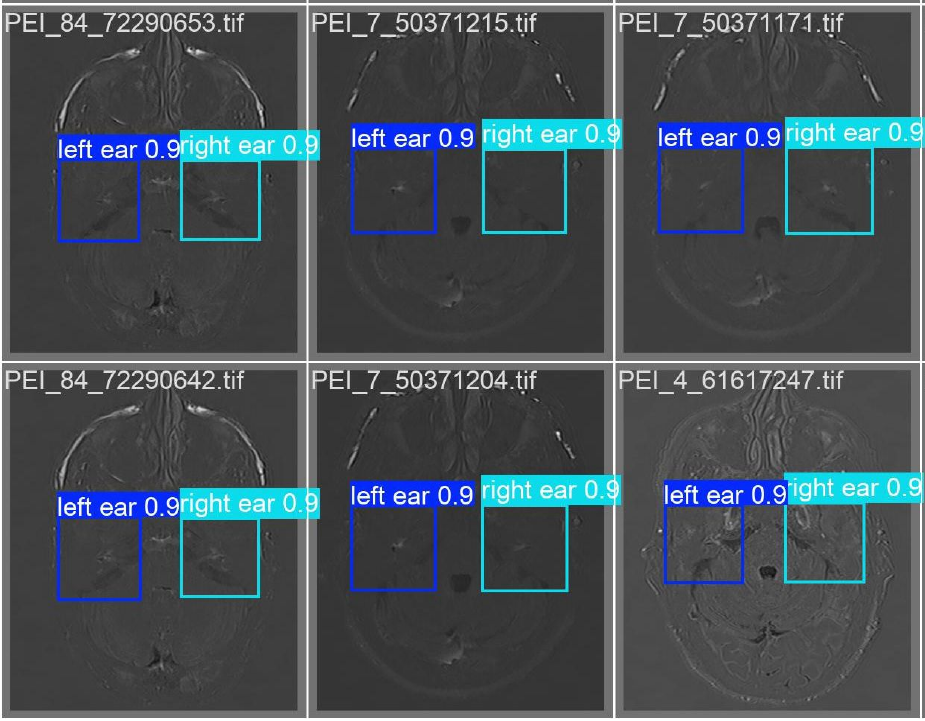}
        %\caption{Representative YOLOv5su detections on the 3D-REAL-IR dataset. Predictions for \textit{left ear} (blue) and \textit{right ear} (cyan) are consistently aligned with manual annotations, confirming robust performance across patients.}
        \label{fig:pei_test_preds}
    \end{subfigure}
    \caption{Representative AuriBox detections across both modalities, showing precise ear localization. (Left) SPACE-MRC: predicted bounding boxes for left (blue) and right (cyan) ears; (Right) REAL-IR: bounding boxes aligned with manual annotations.}
    \label{fig:auribox_preds}
\end{figure}

Qualitative examples (Figs.~\ref{fig:auribox_preds}) show that predicted bounding boxes align closely with manual annotations, confirming accurate spatial localizations across modalities, despite the quality of the acquired images. In fact, the images depicted in Figure \ref{fig:auribox_preds} are the original images seen by the network to perform object detection. 

AuriBox thus provides reliable and computationally efficient localization of the inner-ear regions, ensuring robust cropping and alignment for the downstream segmentation and volumetric analysis stages. 

\subsection{EHMasker performance}

EHMasker performs segmentation of the endolymphatic space within the localized inner-ear regions and constitutes the most critical stage of the pipeline, as small pixel-level errors directly affect ELR. Evaluation focused on three aspects: loss-function selection, architectural and hyperparameter refinements, and binarization threshold optimization. Because annotations were limited to three to six slices per patient, training and validation were conducted on 2D patches rather than full 3D volumes. 

Table~\ref{tab:segmentation-results} summarizes the impact of loss function. Binary Cross-Entropy (BCE) and Focal Loss failed under the severe class imabalance, yielding near-zero Dice and IoU scores. Dice-based objectives substantially improved overlap, and the hybrid BCE+Dice loss provided the most balanced trade-off between precision and recall (Dice = 0.85/0.68 for MEC/REAL, respectively), outperforming both standalone Dice and Tversky losses. this configuration was therefore used in subsequent experiments. 

\begin{table}
\centering
\caption{EHMasker segmentation performance using different loss functions on 3D-SPACE-MRC and 3D-REAL-IR. Dice, IoU, Recall, and final loss are shown; bold values mark the best results.}
\label{tab:segmentation-results}
\small
\renewcommand{\arraystretch}{1.2}
\setlength{\tabcolsep}{6pt}
\begin{tabular}{l l c c c c}
\toprule
\textbf{Dataset} & \textbf{Loss} & \textbf{Dice} & \textbf{IoU} & \textbf{Recall} & \textbf{Loss} \\
\midrule
\multirow{5}{*}{3D-SPACE-MRC}
 & BCE       & 0.02 & 0.00 & 0.00 & 0.044 \\
 & Focal     & 0.02 & 0.00 & 0.00 & 0.003 \\
 & Dice      & 0.84 & 0.75 & 0.85 & 0.008 \\
 & BCE+Dice  & \textbf{0.85} & \textbf{0.76} & \textbf{0.86} & \textbf{0.016} \\
 & Tversky   & 0.85 & 0.75 & 0.86 & 0.006 \\
\midrule
\multirow{5}{*}{3D-REAL-IR}
 & BCE       & 0.13 & 0.00 & 0.00 & 0.044 \\
 & Focal     & 0.13 & 0.00 & 0.00 & 0.003 \\
 & Dice      & 0.61 & 0.47 & 0.72 & 0.026 \\
 & BCE+Dice  & \textbf{0.68} & \textbf{0.52} & \textbf{0.71} & \textbf{0.048} \\
 & Tversky   & 0.68 & 0.48 & 0.60 & 0.017 \\
\bottomrule
\end{tabular}
\end{table}

Progressive architectural refinements (Table~\ref{tab:segmentation-ablation}) demonstrated consistent gains from normalization, LeakyReLU activations, dropout, and data augmentation. The best performance on SPACE-MRC was achieved with a reduced learning rate ($1\text{e}{-4}$, Dice = 0.91, IoU = 0.83, Recall = 0.92), while REAL-IR benefited from a smaller batch size (6), reaching Dice = 0.75. These results highlight the greater segmentation dificulty of REAL-IR data and the benefits of careful hyperparameters tuning from cross-modality generalization. 

\begin{table*}[ht!]
\centering
\caption{Ablation study of EHMaster across architectural and training variants (E1-E10). DSC, IoU, and Recall are reported for both modalities; bold values indicate the best per metric.}
\label{tab:segmentation-ablation}
\small
\renewcommand{\arraystretch}{1.2}
\setlength{\tabcolsep}{6pt}
\begin{tabular}{l c c c c c c}
\toprule
\multirow{2}{*}{\textbf{Experiment}} & \multicolumn{3}{c}{\textbf{3D-SPACE-MRC}} & \multicolumn{3}{c}{\textbf{3D-REAL-IR}} \\
\cmidrule(lr){2-4} \cmidrule(lr){5-7}
 & \textbf{DSC} & \textbf{IoU} & \textbf{Recall} & \textbf{DSC} & \textbf{IoU} & \textbf{Recall} \\
\midrule
E1: Baseline U-Net                        & 0.85 & 0.76 & 0.86 & 0.68 & 0.46 & 0.58 \\
E2: Add Group and   & 0.87 & 0.79 & 0.84 & 0.69 & 0.49 & 0.60 \\
\:\:\:\:\:\:\: Instance Norm. &&&&&&
\\
E3: Add LeakyReLU                         & 0.87 & 0.79 & 0.86 & 0.70 & 0.52 & 0.67 \\
E4: Add Kaiming Init.            & 0.87 & 0.79 & 0.83 & 0.69 & 0.50 & 0.61 \\
E5: Add Dropout  & 0.88 & 0.80 & 0.84 & 0.70 & 0.53 & 0.70 \\
\:\:\:\:\:\:\: (No Kaiming Init.) &&&&&&
\\
E6: Add Dropout & 0.87 & 0.80 & 0.88 & 0.71 & 0.54 & 0.68 \\
\:\:\:\:\:\:\: (No Kaiming Init.) &&&&&&
\\
E7: Add Data Augmentation                 & 0.89 & 0.81 & 0.89 & 0.73 & 0.56 & 0.67 \\
E8: Change learning rate & \textbf{0.91} & \textbf{0.83} & \textbf{0.92} & 0.75 & 0.55 & 0.68 \\
\:\:\:\:\:\:\: (from 1e-3 to 1e-4) &&&&&&
\\
E9: Change batch size       & 0.91 & 0.83 & 0.89 & 0.73 & 0.55 & 0.74 \\
\:\:\:\:\:\:\: (from 16 to 32) &&&&&&
\\
E10: Change batch size to 6               & 0.90 & 0.83 & 0.88 & \textbf{0.75} & \textbf{0.56} & \textbf{0.72} \\
\bottomrule
\end{tabular}
\end{table*}

As the model produces probabilistic maps, we examined the effect of binarization thresholds (Table \ref{tab:segmentation-threshold}). For SPACE-MRC, an intermediate threshold (0.9) yielded the best Dice and Recall (0.91 and 0.92, respectively), whereas for REAL-IR, optimal results were obtained at 0.8 (Dice = 0.75). Excessively low or high thresholds reduced sensitivity or increased noise, confirming the need for modality-specific calibration.

\begin{table}
\centering
\caption{Influence of binarization threshold on EHMasker segmentation for 3D-SPACE-MRC and 3D-REAL.-IR, measured by DSC, IoU, and Recall. Best values are highlighted in bold. }
\label{tab:segmentation-threshold}
\small
\renewcommand{\arraystretch}{1.2}
\setlength{\tabcolsep}{6pt}
\begin{tabular}{l c c c c}
\toprule
\textbf{Dataset} & \textbf{Th} & \textbf{DSC} & \textbf{IoU} & \textbf{Recall} \\
\midrule
\multirow{5}{*}{3D-SPACE-MRC}
 & 0.5  & 0.900 & 0.826  & 0.920 \\
 & 0.7  & 0.902 & 0.829  & 0.934 \\
 & \textbf{0.9}  & \textbf{0.905} & \textbf{0.834} & \textbf{0.923} \\
 & 0.95 & 0.905 & 0.834  & 0.916 \\
 & 0.98 & 0.905 & 0.834  & 0.905 \\
\midrule
\multirow{5}{*}{3D-REAL-IR}
 & 0.5  & 0.719 & 0.563  & 0.731 \\
 & 0.6  & 0.727 & 0.565  & 0.737 \\
 & 0.7  & 0.740 & 0.567  & 0.731 \\
 & \textbf{0.8}  & \textbf{0.745} & \textbf{0.563} & \textbf{0.729} \\
 & 0.9  & 0.739 & 0.560  & 0.707 \\
\bottomrule
\end{tabular}
\end{table}

\begin{figure}[ht!]
    \centering
    \includegraphics[width=0.7\linewidth]{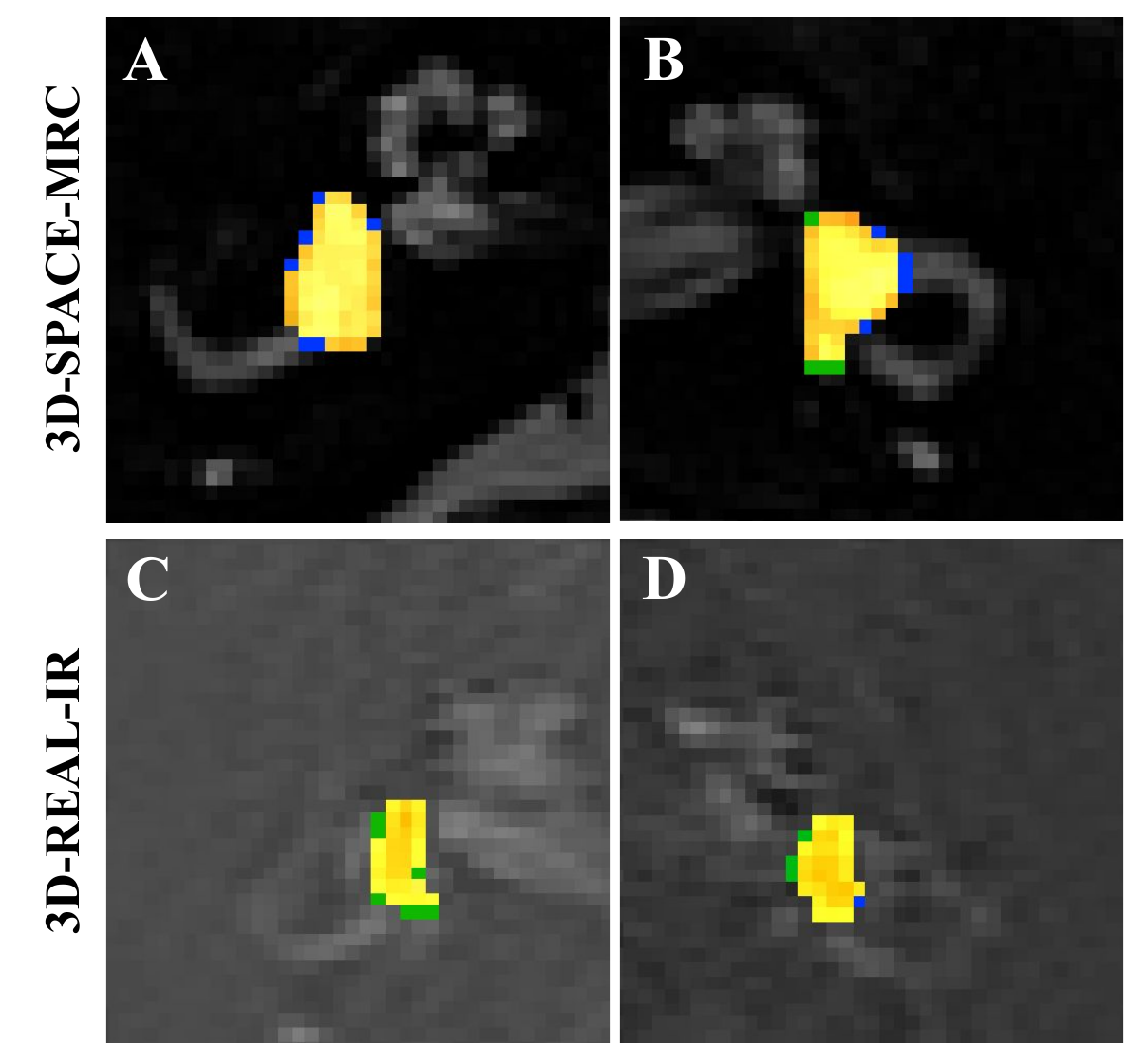}
    \caption{Qualitative segmentation examples for both modalities. Panels A–B: 3D-SPACE-MRC; Panels C–D: 3D-REAL-IR. The model prediction (green) and the ground-truth annotation (blue) are overlaid on the MRI slice; the overlap region is shown in yellow. These examples illustrate the close agreement between predicted masks and expert annotations across modalities.}
    \label{fig:ehmasker_qualitative}
\end{figure}

Representative examples in Fig. \ref{fig:ehmasker_qualitative} illustrate the close alignment between predictions and manual annotations. Most errors occurred in low-contrast slices or regions affected by partial-volume artifacts. Overall, EHMasker achieved robust performance, particularly for SPACE-MRC (Dice > 0.9), and maintained reliable sensitivity in REAL-IR, validating the U-Net with BCE+Dice loss and the adopted training strategy as an effective approach for automated segmentation of the endolymphatic space.

\subsection{Volumetric evaluation and ELR Quantification}

The ELR represents the volumetric proportion between the endolymphatic and vestibular cavities. Its computation requires complete 3D segmentation of both structures. Because manual annotations were available for only 3-6 slices per patient, these partial labels were unsuitable as volumetric ground truth. Consequently, ELR could only be estimated after applying the full OREHAS pipeline, which performs slice-wise segmentation across the entire MRI volume and reconstructs the complete 3D ear anatomy. This design aligns with the intended clinical use, where the pipeline must operate fully automatically on whole scans.

Two datasets were analyzed: (i) the held-out test patients from model development and (ii) an independent cohort of five fully annotated cases provided by clinical collaborators for external validation. 

\subsubsection{Test patients}

The OREHAS pipeline was first evaluated on 19 held-out patients not seen during training. Because ELR is a volumetric metric, all reported results correspond to the complete end-to-end pipeline—combining classification, detection, segmentation, and quantification—to reflect real deployment performance.

Volumetric reference values were obtained with the syngo.via clinical software. As this tool uses partial manual annotations with proprietary interpolation, it does not provide true slice-level ground truth; full manual annotations were reserved for the external validation cohort described below.

Across the test set, vestibular volumes measured on SPACE-MRC showed close agreement between syngo.via ($0.083 \pm 0.013$ cm³) and OREHAS ($0.084 \pm 0.010$ cm³). Endolymphatic volumes measured on REAL-IR were lower with OREHAS ($0.019 \pm 0.008$ cm³) compared with syngo.via ($0.043 \pm 0.020$ cm³). Consequently, OREHAS yielded lower ELR values—$20.2\%$ (right) and $25.7\%$ (left)—relative to the reference ($46.0\%$ and $57.6\%$). Table~\ref{tab:elr_summary_test_by_ear} summarizes these findings, showing consistent vestibular measurements but systematically lower endolymphatic volumes and ELR values with OREHAS. 

\begin{table}
\centering
\caption{Per-ear comparison of vestibular and endolymph volumes and ELR (mean $\pm$ std, $N{=}19$) between syngo.via and OREHAS for 3D-SPACE-MRC and 3D-REAL-IR.}
\label{tab:elr_summary_test_by_ear}
\small
\renewcommand{\arraystretch}{1.2}
\setlength{\tabcolsep}{6pt}
\begin{tabular}{l l c c}
\toprule
\textbf{Ear} & \textbf{Measurement} & \textbf{Reference (syngo.via)} & \textbf{Automatic (OREHAS)} \\
\midrule
\multirow{3}{*}{Right}
 & Vestibule (cm$^3$) & $0.084 \pm 0.012$ & $0.086 \pm 0.011$ \\
 & Endolymph (cm$^3$) & $0.038 \pm 0.018$ & $0.017 \pm 0.008$ \\
 & ELR (\%)           & $46.0 \pm 22.2$   & $20.2 \pm 9.0$ \\
\midrule
\multirow{3}{*}{Left}
 & Vestibule (cm$^3$) & $0.083 \pm 0.015$ & $0.082 \pm 0.010$ \\
 & Endolymph (cm$^3$) & $0.047 \pm 0.022$ & $0.021 \pm 0.008$ \\
 & ELR (\%)           & $57.6 \pm 26.1$   & $25.7 \pm 11.6$ \\
\bottomrule
\end{tabular}
\end{table}

Figure \ref{fig:volumes_by_ear_reference_vs_auto} compares vestibular and endolymphatic volumes for both ears. Vestibular measurements were highly consistent between methods, whereas endolymph volumes estimated by OREHAS were systematically smaller, leading to lower ELR values. This systematic offset and its clinical implications are analyzed further in the discussion.

\subsubsection{External Validation Patients}\label{subsec:external-val}
As described in Section~\ref{subsec:pipeline-integration}, five additional patients from CUN were used for external validation. Unlike syngo.via—routinely employed in clinical practice but operating as a black box—these cases were manually annotated at the pixel level, providing transparent and reliable ground truth for the full MRI volumes. This enabled direct comparison of OREHAS predictions with both manual and syngo.via-derived ELR values.

\begin{figure}[ht!]
    \centering
    \includegraphics[width=0.75\linewidth]{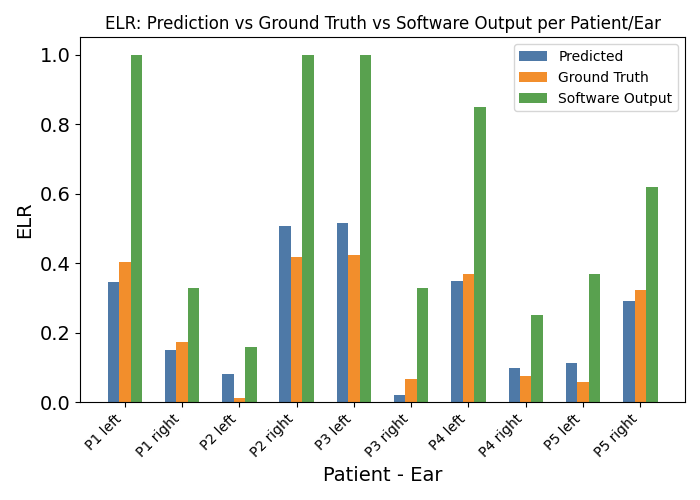}
    \caption{Comparison of Endolymph-to-Vestibule Ratio (ELR) values for the five fully annotated patients. Results are shown per ear for three sources: OREHAS predictions (blue), manual ground truth (orange), and syngo.via software outputs (green).}
    \label{fig:ELR-5pat-comparison}
\end{figure}

Figure~\ref{fig:ELR-5pat-comparison} compares ELR estimates per patient and ear across the three methods. OREHAS closely matched the manual ground truth, while syngo.via displayed higher variability and a consistent overestimation of the endolymphatic ratio. 
 
Table~\ref{tab:vsi_summary} details quantitative comparisons with the manual annotations using the volumetric similarity index (VSI). In SPACE–MRC (vestibular cavity), OREHAS and syngo.via achieved comparable agreement with manual ground truth ($86.1\% \pm 0.04$ and $88.8\% \pm 0.04$, respectively). For REAL–IR (endolymphatic space), OREHAS substantially outperformed syngo.via ($74.3\% \pm 0.22$ vs. $42.5\% \pm 0.17$), confirming the robustness of the proposed segmentation and volume reconstruction.

\begin{table}
\centering
\caption{Comparison of OREHAS and syngo.via volumes with manual ground truth for the external validation cohort (N=5), including volumetric similarity index (VSI).}
\label{tab:vsi_summary}
\small
\renewcommand{\arraystretch}{1.2}
\setlength{\tabcolsep}{4pt}
\begin{tabular}{l c c c c c}
\toprule
 & \multicolumn{3}{c}{\textbf{Volume (mm$^3$)}} & \multicolumn{2}{c}{\textbf{VSI (\%)}} \\
\cmidrule(lr){2-4} \cmidrule(lr){5-6}
\textbf{Dataset} & \textbf{Pred.} & \textbf{Syngo.via} & \textbf{GT} & \textbf{OREHAS} & \textbf{Syngo.via} \\
\midrule
3D-SPACE--MRC & $92.51 \pm 16.89$ & $87.00 \pm 11.60$ & $69.54 \pm 9.68$  & $86.1 \pm 0.04$ & $88.8 \pm 0.04$ \\
3D-REAL--IR   & $22.86 \pm 18.43$ & $51.40 \pm 32.28$ & $16.34 \pm 12.61$ & $74.3 \pm 0.22$ & $42.5 \pm 0.17$ \\
\bottomrule
\end{tabular}
\end{table}

\section{Discussion}
\label{sec:disc}

\subsection{Key Findings and Clinical Implications}
OREHAS achieved robust segmentation and volumetric quantification despite limited supervision, demonstrating the feasibility of fully automatic EH analysis from routine MRI. Trained with only 3–6 annotated slices per patient, the EHMasker module generalized effectively to full 3D volumes, reaching Dice scores above 0.9 for SPACE–MRC and 0.75 for REAL–IR. The external validation cohort (Section~\ref{subsec:external-val}) provided the most relevant confirmation, where OREHAS predictions closely matched full manual ground truth (VSI = 74.3\% for REAL–IR) and clearly outperformed the clinical syngo.via software (VSI = 42.5\%), which systematically overestimated endolymphatic volumes. These results highlight that current diagnostic thresholds for hydrops severity (e.g., ELR $>$60\% for significant and $<$30\% for non-significant) \cite{paper_medicos}—originally derived from syngo.via outputs—may require recalibration based on accurate, fully volumetric measurements.

Two factors likely underlie these differences. First, the endolymphatic structures are extremely small, often spanning only 3–10 pixels per slice, so even a single-pixel boundary mismatch can induce large volumetric errors. This contrasts with prior segmentation benchmarks reporting very high Dice values (96\% for craniofacial bones, 98\% for the mandible, 78.8\% for the cranial base) \cite{sun2025automatic}, where larger and better-defined structures dominate. Similar challenges have been noted by Ding et al. \cite{ding2023self}, who achieved high Dice for the entire bony labyrinth but emphasized the difficulty of segmenting smaller ear substructures, suggesting that the endolymph is an order of magnitude harder to delineate. Second, syngo.via operates as a proprietary black box: clinicians outline a few slices, and the software interpolates across the volume using undisclosed algorithms. This lack of transparency may bias volumes upward, as seen in our comparisons. Furthermore, whereas OREHAS computes volumes with millimeter precision, syngo.via reports only cubic-centimeter values, potentially introducing rounding errors and loss of precision.

From a clinical standpoint, these findings have significant implications. Diagnostic ELR thresholds currently rely on syngo.via-derived values, which our results indicate may not reflect true endolymphatic volumes. The occurrence of ELR values exceeding 100\%—biologically implausible if the endolymph is contained within the vestibule—underscores the need to revise these cut-offs using reproducible, physically consistent measurements. By providing an open, and fully automated volumetric pipeline, OREHAS offers a more accurate and reproducible foundation for future diagnostic guidelines in endolymphatic hydrops assessment.

\subsection{Comparison with prior semi-and fully automated methods}
Previous clinical workflows for ELR quantification have been predominantly semi-automatic, relying on expert delineation of a few slices followed by volumetric interpolation through proprietary software—an accurate but time-consuming and operator-dependent process \cite{paper_medicos, Naganawa}.

Among automated solutions, INHEARIT and INHEARIT-v2 introduced CNN-based segmentation and surface-derived ELR estimation from MRC/FLAIR/REAL-IR sequences \cite{Cho2020, Park2021}. However, these methods remain focused on segmentation of preselected slices rather than full 3D processing within an integrated pipeline. Other strategies emphasize 3D segmentation of the entire inner ear, often requiring substantial data and computational resources \cite{Vaidyanathan, Yoo}, or rely on non-DL thresholding approaches that omit ELR computation altogether~\cite{VOLT}.

In contrast, OREHAS provides a fully automatic, end-to-end framework encompassing slice classification, inner-ear localization, and sequence-specific segmentation directly on standard 3D-SPACE-MRC and 3D-REAL-IR volumes. This design minimizes manual intervention, standardizes every processing stage, and outputs volumetric ELR per ear—addressing the practical limitations of prior semi-automatic and segmentation-only methods \cite{paper_medicos, Cho2020, Park2021, Yoo}.

\subsection{Limitations (dataset size, variability, annotation bias)}
This study was based on a relatively small cohort (arund 15 patients), which limits statistical power and generalizability. Nonetheless, each subject contributed two MRI sequences (3D-SPACE-MRC and 3D-REAL-IR) and numerous slices, yielding substantial data volume and computational demand for end-to-end evaluation.

Variability may arise both across patients—due to anatomical differences, pathology severity, and acquisition settings—and within patients, owing to contrast, motion, or slice-thickness differences between sequences. Annotation bias is also possible (e.g., single-rater boundaries or thresholding inconsistencies), although its effect is mitigated by the large number of labeled slices per case and standardized annotation protocols.

Finally, as in any multi-stage pipeline, errors may propagate across modules. While the modular architecture supports debugging and component replacement, broader multi-center validation will be essential to assess robustness under larger domain shifts.

\subsection{Future directions}
First, expanded annotations will enable stronger volumetric supervision and more rigorous benchmarking beyond slice-level labels. Second, clinical threshold calibration will be pursued to relate ELR outputs to decision-ready categories used in practice (e.g., ELR ranges for hydrops severity), including per-sequence operating points optimized for robustness. Third, we plan to deploy OREHAS as a containerized syngo.via OpenApp\footnote{https://www.siemens-healthineers.com/digital-health-solutions/syngovia-openapps} for research-grade use under clinical supervision.

The application will ingest DICOM studies, process 3D-SPACE-MRC and/or 3D-REAL-IR volumes, and return DICOM overlays with optional CSV summaries (slice selection, ROIs, ELR per ear) directly accessible through the syngo.via interface. Even at the current stage, our aim is to install the tool on-site for retrospective analyses, annotation support, and consistency checks—without diagnostic intent—while performance continues to be refined using larger, multi-center datasets.

\section{Conclusions}
\label{sec:conc}
We introduced OREHAS, a fully automatic and open pipeline for volumetric quantification of endolymphatic hydrops (EH) from routine 3D-SPACE-MRC and 3D-REAL-IR MRI. The system integrates slice classification, inner-ear localization, and sequence-specific segmentation to compute per-ear ELR with reproducible accuracy.

Trained with sparse annotations, OREHAS generalized across full volumes, achieving high Dice scores (0.9 for SPACE-MRC, 0.75 for REAL-IR) and strong agreement with manual ground truth while outperforming the proprietary syngo.via software, which systematically overestimated endolymphatic volumes. These findings suggest that currently used clinical thresholds—derived from syngo.via measurements—may require recalibration based on physically consistent, fully volumetric data.

OREHAS establishes a transparent, modular, and auditable workflow that standardizes each processing step, reduces operator dependency, and supports reproducible analysis of EH. Integration as a syngo.via OpenApp will facilitate on-site clinical use for retrospective and research applications. Future efforts will expand annotation coverage, refine diagnostic thresholds, and pursue multi-center validation to consolidate an open benchmark for automatic inner-ear analysis.

\section*{CRediT authorship contribution statement}
Caterina Fuster-Barceló: Conceptualización, Data Curation, Formal analysis, Investigation, Methodology, Software, Supervision, Validation, Writing – original draft, Writing – review and editing. Claudia Castrillón: Data Curation, Software, Validation, Writing – review and editing. Laura Rodrigo Muñoz: Data Curation, Software, Validation, Writing – review and editing. Victor Manuel Vega-Suárez: Formal analysis, Investigation, Methodology, Validation, Writing – review and editing. Nicolás Pérez-Fernández: Investigation, Project administration, Supervision, Writing – review and editing; Gorka Bastarrika: Investigation, Writing – review and editing; Arrate Muñoz-Barrutia: Conceptualización, Formal analysis, Investigation, Methodology, Project administration, Supervision, Writing – review and editing.

\section*{Declaration of competing interest}
The authors declare that they have no known competing financial interests or personal relationships that could have appeared to influence the work reported in this paper. 

\section*{Ethics statement}
The studies involving humans were approved by Research Ethics Committee of the University of Navarra (project number 2021.199). The studies were conducted in accordance with the local legislation and institutional requirements. The participants provided their written informed consent to participate in this study. Written informed consent was obtained from the individual(s) for the publication of any potentially identifiable images or data included in this article.

\section*{Acknowledgements}
The authors want to thank Gloria del Rocío Muñoz for her contribution of this work on the early stages. This work was partially supported by the European Union’s Horizon Europe research and innovation program under grant agreement number 101057970 (AI4Life project) awarded to A.M.-B.; also under grants PID2023-152631OB-I00 and AIA2025-164165-C41 by the Ministerio de Ciencia, Innovación y Universidades, Agencia Estatal de Investigación (MCIN/AEI/10.13039/501100011033/), co-financed by European Regional Development Fund (ERDF), ’A way of making Europe’. This study was funded by the Instituto de Salud Carlos III through projects PI19/00414 and PI24/00175, co-funded by the European Regional Development Fund/European Social Fund, “A way to make Europe”. C.F.B. acknowledges funding from the University of Zurich. 

During the preparation of this work the authors used ChatGPT in order to check grammar and spelling. After using this tool, the authors reviewed and edited the content as needed and take full responsibility for the content of the published article.

\appendix
\section{Losses Ablation Study}
\label{app:losses}

As summarized in Table~\ref{tab:segmentation-results}, we performed an ablation study over several loss functions for training the segmentation network: Binary Cross-Entropy (BCE), Focal loss, soft Dice loss, a combined BCE+Dice loss, and the Tversky loss. For completeness, we provide here the definitions used in the binary segmentation setting.

Let $\Omega$ denote the set of pixels (or voxels), $y_i \in \{0,1\}$ the ground-truth label at pixel $i$, and $p_i \in [0,1]$ the predicted probability (after the sigmoid layer). We also use a small constant $\varepsilon > 0$ for numerical stability where indicated.

\paragraph{\textbf{Binary Cross-Entropy (BCE)}}
The BCE loss is given by
\begin{equation}
    L_{\mathrm{BCE}}(p, y)
    = - \frac{1}{|\Omega|}
    \sum_{i \in \Omega}
    \left[
        y_i \log(p_i)
        + (1 - y_i)\log(1 - p_i)
    \right].
\end{equation}

\paragraph{\textbf{Focal loss}}
The Focal loss down-weights easy examples and focuses training on hard, misclassified pixels. With parameters $\alpha \in [0,1]$ and $\gamma \geq 0$, it is defined as
\begin{equation}
    L_{\mathrm{Focal}}(p, y)
    = - \frac{1}{|\Omega|}
    \sum_{i \in \Omega}
    \Big[
        \alpha \, (1 - p_i)^{\gamma} \, y_i \log(p_i)
        + (1 - \alpha) \, p_i^{\gamma} \, (1 - y_i)\log(1 - p_i)
    \Big].
\end{equation}

\paragraph{\textbf{Dice loss}}
The Dice coefficient is defined as
\begin{equation}
    \mathrm{Dice}(p, y)
    = \frac{2 \sum_{i \in \Omega} p_i y_i + \varepsilon}
           {\sum_{i \in \Omega} p_i + \sum_{i \in \Omega} y_i + \varepsilon},
\end{equation}
and the corresponding Dice loss is
\begin{equation}
    L_{\mathrm{Dice}}(p, y)
    = 1 - \mathrm{Dice}(p, y).
\end{equation}

\paragraph{\textbf{Combined BCE+Dice loss}}
The combined loss used in the ablation study is the sum of the BCE and soft Dice losses:
\begin{equation}
    L_{\mathrm{BCE+Dice}}(p, y)
    = L_{\mathrm{BCE}}(p, y)
    + L_{\mathrm{Dice}}(p, y).
\end{equation}

\paragraph{\textbf{Tversky loss}}
The Tversky index generalizes the Dice coefficient by weighting false positives and false negatives differently. With parameters $\alpha_T, \beta_T \geq 0$, it is defined as
\begin{equation}
    \mathrm{TI}(p, y)
    = \frac{\sum_{i \in \Omega} p_i y_i + \varepsilon}
           {\sum_{i \in \Omega} p_i y_i
            + \alpha_T \sum_{i \in \Omega} (1 - y_i)p_i
            + \beta_T \sum_{i \in \Omega} y_i (1 - p_i)
            + \varepsilon},
\end{equation}
and the corresponding Tversky loss is
\begin{equation}
    L_{\mathrm{Tversky}}(p, y)
    = 1 - \mathrm{TI}(p, y).
\end{equation}

The quantitative results in Table~\ref{tab:segmentation-results} (BCE, Focal, Dice, BCE+Dice, and Tversky) correspond to models trained with each of these loss functions separately.

\section{Supplementary Volume Analysis}
\label{app:volumes}

\begin{figure}
    \centering
    \includegraphics[width=\linewidth]{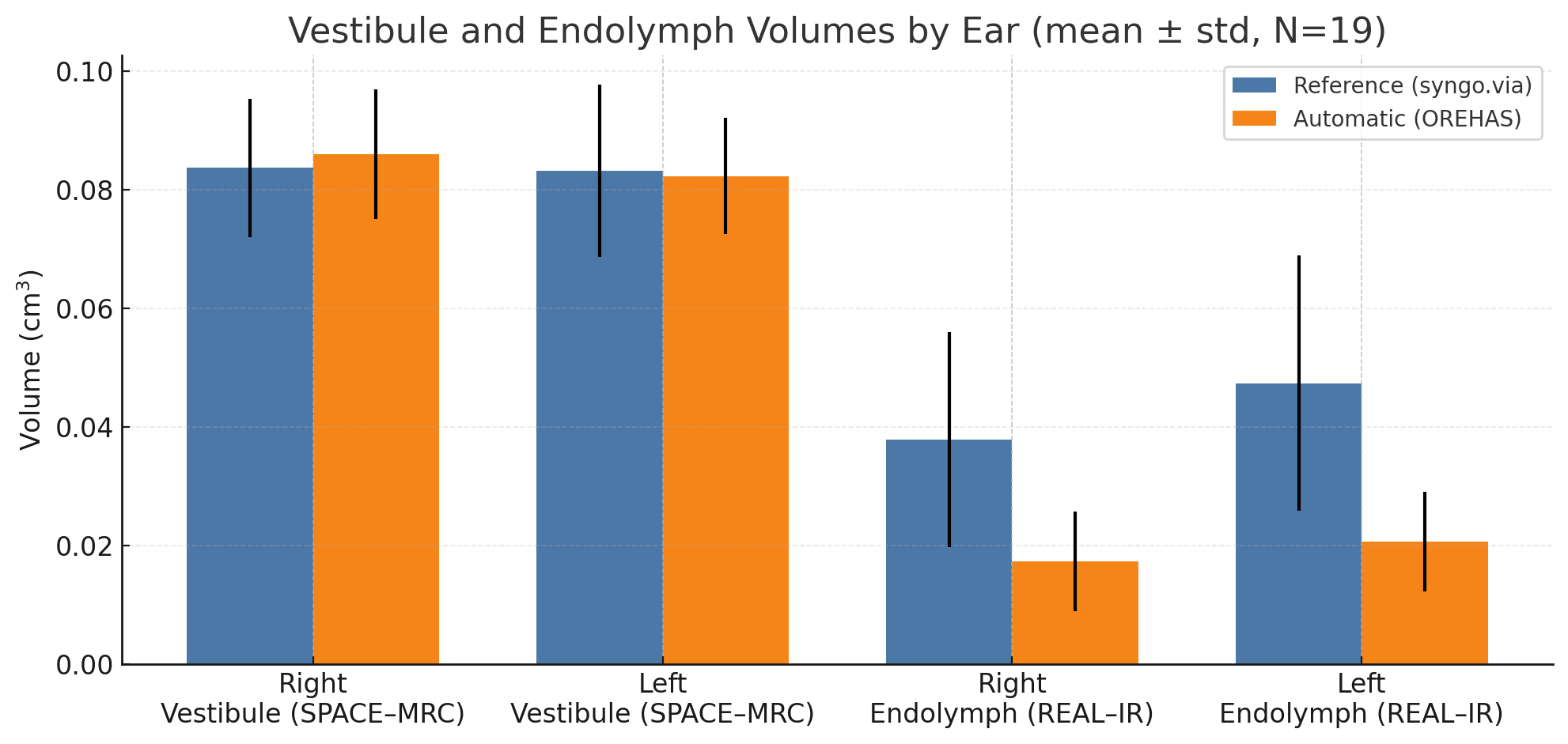}
    \caption{Comparison of vestibular (SPACE–MRC) and endolymphatic (REAL–IR) volumes between syngo.via and OREHAS (mean ± std, $N{=}19$). OREHAS shows comparable vestibular estimates but lower endolymphatic volumes.}
    \label{fig:volumes_by_ear_reference_vs_auto}
\end{figure}

As a complement to the volumetric results reported in Table~\ref{tab:elr_summary_test_by_ear}, we additionally provide a graphical comparison of vestibular (SPACE--MRC) and endolymphatic (REAL--IR) volumes between the clinical reference (syngo.via) and OREHAS (Fig.~\ref{fig:volumes_by_ear_reference_vs_auto}). This visualization highlights the agreement in vestibular volumes and the systematically lower endolymphatic volumes obtained with OREHAS.

%%       \cite{<label>} ==> [1]




\bibliographystyle{elsarticle-harv} 
\bibliography{bibliography}

\end{document}